\documentclass{article}
\usepackage{float}
\usepackage{float}

\usepackage[sglblindworkshop,final]{neurips_2025}
\usepackage{multirow}   
\usepackage{tabularx}   
\usepackage{array}      
\usepackage{multirow}
\usepackage{tabularx}
\usepackage{booktabs}   
\usepackage{adjustbox}  
\usepackage{tcolorbox}
\tcbuselibrary{most}

\usepackage{enumitem}

\usepackage{setspace}

\usepackage[utf8]{inputenc} 
\usepackage[T1]{fontenc}    
\usepackage{hyperref}       
\usepackage{url}            
\usepackage{booktabs}       
\usepackage{amsfonts}       
\usepackage{nicefrac}       
\usepackage{microtype}      
\usepackage{xcolor}         

\title{AuditCopilot: Leveraging LLMs for Fraud Detection in Double-Entry Bookkeeping}
\workshoptitle{Generative AI in Finance}

\author{
  Md Abdul Kadir\textsuperscript{1,2} \quad
  Sai Suresh Macharla Vasu\textsuperscript{1,3}\thanks{Equal contribution.} \quad
  Sidharth S.\ Nair\textsuperscript{1,3}\footnotemark[1] \quad
  Daniel Sonntag\textsuperscript{1,2} \\
  \textsuperscript{1}\,German Research Center for Artificial Intelligence (DFKI), Saarbrücken, Germany \\
  \textsuperscript{2}\,Oldenburg University, Oldenburg, Germany \\
  \textsuperscript{3}\,Saarland University, Saarbrücken, Germany \\
  \texttt{\{abdul.kadir, sai\_suresh.macharla\_vasu, sidharth.nair, daniel.sonntag\}@dfki.de}
}

\usepackage[T1]{fontenc}
\usepackage[utf8]{inputenc}
\usepackage{graphicx}
\usepackage{booktabs}
\usepackage{amsmath,amssymb}
\usepackage{tikz}
\usetikzlibrary{arrows.meta}
\usetikzlibrary{positioning}

\begin{document}
\maketitle

\begin{abstract}
\vspace{-5pt}
Auditors rely on Journal Entry Tests (JETs) to detect anomalies in tax-related ledger records, but rule-based methods generate overwhelming false positives and struggle with subtle irregularities. We investigate whether large language models (LLMs) can serve as anomaly detectors in double-entry bookkeeping. Benchmarking SoTA LLMs such as LLaMA and Gemma on both synthetic and real-world anonymized ledgers, we compare them against JETs and machine learning baselines. Our results show that LLMs consistently outperform traditional rule-based JETs and classical ML baselines, while also providing natural-language explanations that enhance interpretability. These results highlight the potential of \textbf{AI-augmented auditing}, where human auditors collaborate with foundation models to strengthen financial integrity.

\end{abstract}

\vspace{-15pt} 
\section{Introduction}
\vspace{-10pt} 
Financial auditors play a critical role in detecting unusual transactions and anomalies that could indicate errors, fraud, or tax evasion in a company’s books. Ensuring that such irregularities are identified is vital for maintaining trust in financial statements~\cite{wang2025automating}. However, auditing large volumes of accounting data is extremely challenging; current manual and rule-based processes are often inefficient and error-prone, meaning auditors can miss important red flags. Standard Journal Entry Tests (JETs)~\cite{droste2024jet}, which rely on predefined rules and known patterns, tend to flag a huge number of transactions. These full population tests catch only anticipated anomalies~\cite{herreros2023applied} and flood auditors with alerts, many of which are false positives requiring time-consuming review. In practice, auditors still rely heavily on sampling and human labor~\cite{wang2025automating}, which risks overlooking subtle irregularities in today’s era of massive, complex datasets~\cite{herreros2023applied,wang2025automating}. This gap between growing data complexity and the limits of traditional audit techniques underscores the need for more intelligent, automated anomaly detection tools in auditing.

A journal entry, in accounting terms, represents the basic record of a financial transaction, typically involving a debit and a credit entry across one or more accounts. Each entry is enriched with attributes such as transaction date, posting ID, currency, user responsible for entry, tax rates, amounts, and free-text descriptions. These heterogeneous features, ranging from numerical to categorical to textual, make journal entries a natural testbed for anomaly detection systems operating in multi-type data environments. Both the synthetic and anonymized datasets share a common feature space that includes these core attributes, facilitating consistent modeling and evaluation across domains.

In response, researchers have explored machine learning (ML) to enhance anomaly detection in accounting data. Unsupervised algorithms, such as clustering and Isolation Forest, can sift through journal entries to pinpoint outliers that deviate from normal patterns~\cite{herreros2023applied}. Recent work on synthetic general ledger data demonstrated that these methods can improve detection performance beyond rule-based JETs~\cite{pub15550}. Still, there are limitations: traditional ML models typically operate on structured features and may lack the contextual understanding needed to differentiate truly risky anomalies from benign outliers.

Meanwhile, the advent of \textbf{Large Language Models (LLMs)} has opened up new possibilities for auditing. These foundation models, exemplified by GPT-4 and similar, possess powerful natural language understanding and reasoning capabilities. Early studies suggest that LLMs could serve as AI auditors or co-pilots, working alongside humans to analyze financial data~\cite{gu2023aicopilot}. Recent advances also indicate that LLMs can be leveraged for \textbf{anomaly detection}, such as SigLLM for time-series~\cite{mit2024sigllm} and AnoLLM for tabular records~\cite{tsai2025anollm}, both showing competitive performance without heavy feature engineering.

Of course, deploying LLMs in high-stakes audit applications must be done carefully. By nature, LLMs can sometimes produce inconsistent answers, show biases, or hallucinate. Recognizing this, the community has begun to audit the auditors with approaches such as AuditLLM~\cite{amirizaniani2024auditllm} and LLMAuditor~\cite{amirizaniani2024llmauditor}. These highlight that while LLMs are powerful, governance and oversight are necessary when using them as decision-support tools in domains like auditing.

In this paper, we leverage general-purpose LLMs for anomaly detection on tax-related ledger data, combining data-driven detection with natural-language explainability using structured input output schema making a scalable method. We evaluate this on two datasets: a large real-world anonymized general ledger obtained from project partners and a synthetic dataset from~\cite{pub15550}. By testing on both, we ensure our method works for practical, messy data and benchmark scenarios. Our results show that LLMs can achieve detection performance comparable to classic ML algorithms, while simultaneously providing rich, interpretable explanations that can assist auditors. This represents an encouraging step toward \textbf{AI-augmented auditing}, where human auditors and AI systems collaborate to ensure financial integrity.


\vspace{-10pt}

\section{Method}
\vspace{-5pt}
\vspace{-5pt}
We use prompt tuning of a large language model to decide whether a posting ID (journal-entry (JE) grouping) is fraudulent or normal and to emit a concise explanation. Model weights are never updated; behaviour is controlled by an instruction prompt and a structured input schema. The method operates under weak/no supervision (analogous to Isolation Forest), while natively ingesting heterogeneous numeric, categorical, and textual JE fields without heavy feature engineering. We find the benefits of this method to be multi-fold, grounding explanations are key to establishing trust in financial AI models for non-technical stakeholders, and prompt tuning is a fairly straightforward modification in simple natural language, leading to behaviour modifications in LLMs' decision making, hence giving more control to end-users in adjusting the model's behaviour for specifc use cases.
\vspace{-10pt}
\paragraph{Addition of contextual information} Contextual information is important in delicate decision-making scenarios such as financial auditing, even for human auditors. Motivated by this fact, we augment the LLM prompt by adding summary statistics and percentile information for numerical features such as transaction amounts, and adding frequency of occurrences for categorical features, such as user ID of accountants. As described earlier, when ingesting fine-grained data per transaction instead of posting ID, we find performance gains in our method when adding global dataset statistics, as well as grounding of the explanations with reduced hallucinations, this is corroborated by AD‑LLM \cite{yang2024ad} shows prompt augmentation with semantic context improves discriminatory performance.
Additionally, we add more signal to the prompt by adding the decision of an Isolation Forest \cite{isolation_forest} classifier into the prompt. Empirically, we found this improved performance wrt the number of False Positives (FP) registered by our method, effectively reducing the burden of their reevaluation by human auditors, saving time and increasing efficiency.
\vspace{-15pt}
\paragraph{Prompt specifics} Prompts are broken down into system and transaction-specific parts. System prompts include general guidelines as well as the contextual information discussed above.  Each instance (posting ID/transaction) is provided as a compact JSON-like record with multi-type fields (date, posting\_id, currency, user, tax rates, amounts, account codes, memo text). The model returns a strict JSON object with \texttt{anomaly} $\in\{0,1\}$ and an \texttt{explanation}.

\vspace{-10pt}
\section{Dataset and Experiments}
\vspace{-5pt}

\vspace{-5pt}
The task is framed as \textbf{unsupervised fraud detection in multi-type journal entry data}. Labels for fraudulent activity are scarce and often unavailable in practice, making supervised learning infeasible at scale. Instead, we aim to detect anomalous entries by modeling irregular patterns across structured and unstructured fields to reflect real-world auditing conditions.  
\vspace{-10pt}
\paragraph{Synthetic JE dataset}This is a \textbf{simulated accounting dataset} introduced by ~\cite{pub15550}, designed to emulate the day-to-day book-keeping practices of a medium-sized enterprise. Features in each entry include a unique posting ID, date and time of posting and of the transaction, Credit/Debit (CD) flags, tax rate, Account ID etc. 
\vspace{-10pt}
\paragraph{Anonymized JE dataset} We benchmark our method on real-world anonymized accounting data from our industry partner. Unlike typical double-entry bookkeeping systems where posting IDs link balanced debit-credit cycles, this dataset lacks such identifiers due to anonymization, breaking the usual transactional structure. We therefore treat it as a transaction-level fraud detection problem. To compensate for missing context, we enrich the LLM prompt with bookkeeping heuristics, guidelines, and summary statistics, which proves essential to outperform traditional ML baselines.


Privacy is an important aspect when dealing with confidential information such as accounting data, hence preference was given to open weight LLMs (LLaMa \cite{llama3.1_2025}, Gemma \cite{team2024gemma}) over proprietary LLM API endpoints such as (GPT 4o \cite{openai_gpt4o_2024}) when benchmarking our method. 

Real world (anonymized) and synthetic JE data has been used for our tests. For the real-world accounting data, due to the non-availablity of labels, pseudo-labeling was done purely for evaluations by employing Journal Entry Testing (JET) practices verified by the client involved in the project. JET is a rule-based method used in practice to filter out outliers for further scrutiny by auditors. Control over the anomaly rate is possible in the case of Synthetic data, where we present detailed results for a randomly sampled subset of 5000 posting ID's with a 1\% anomaly rate i.e. 50 anomalous entries. For synthetic data, post pseudo labeling with JET, we get $\sim$ 6\% anomaly rate.
\vspace{-10pt}
\section{Results and Discussion}
\vspace{-10pt}
Our method outperforms the baselines provided on the Synthetic data \cite{pub15550} as well as on a real-world anonymised dataset. We use Isolation forest \cite{isolation_forest} as our baseline, due to its theoretical advantages in modeling tabular data \cite{grinsztajn2022tree}, additionally, it is an unsupervised anomaly detection method, hence aligning with the proposed task in Section 3.
\vspace{-10pt}
\paragraph{Synthetic dataset.} 
Table~\ref{tab:synthetic-all-methods} summarizes results on the synthetic benchmark from~\cite{pub15550}. 
The traditional JET baseline achieves reasonable recall (0.90) but suffers from very high false positives (FP=942), highlighting its limited precision in practice. 
Isolation Forest, substantially improves over JET with higher precision (0.61) and near-perfect recall (0.98), reducing false positives by an order of magnitude (FP=169). 

Among the LLMs, Mistral-8B \cite{ministral8b_2024} delivers the strongest overall performance, reaching the best F1 score (0.94), the highest precision (0.90), and the lowest false positives (FP=12) while maintaining very high recall (0.98). 
Gemma-7B achieves the highest recall (0.99, FN=0), but at a moderate precision cost (0.71). 
GPT-5-mini \cite{openai_gpt5mini_2025} also attains perfect recall (FN=0) but still produces several hundred false positives (FP=466). 
Other models, such as Llama-3.1-8B and Gemma-2B, show balanced but weaker trade-offs. 
Overall, these results indicate that modern LLMs, particularly Mistral-8B, can match or surpass traditional ML baselines on synthetic accounting data, achieving both low error rates and interpretable natural-language outputs.

\vspace{-10pt}
\paragraph{Prompt Ablation results} Table 2 shows that the full \textit{AuditCopilot} prompt yields the most balanced performance, with Gemma-2B achieving the best recall (0.98, FN=6) and Gemma-7B the highest precision (0.89, FP=32). Removing \textit{Isolation Forest} drastically increases false positives (e.g., Gemma-7B: 32$\rightarrow$1973), while removing \textit{Statistics} collapses recall across all models (e.g., Gemma-2B: FN 6$\rightarrow$306). These findings indicate that Statistics are key for recall, while Isolation Forests are crucial for precision, and both are necessary for reliable anomaly detection.

\vspace{-5pt}
\begin{table}[h]
\centering
\caption{Anomaly detection results on the Synthetic dataset \cite{pub15550}, using the prompt template in Fig. \ref{fig:prompt-synthetic}. Best value per column in \textbf{bold}; lowest for FP/FN.}
\label{tab:synthetic-all-methods}
\renewcommand{\arraystretch}{1.3}
\resizebox{\textwidth}{!}{%
\begin{tabular}{lllrrrrrrr} 
\hline
Dataset & Method Group & Method &  Precision $\uparrow$ & Recall $\uparrow$ & F1 $\uparrow$ & TP $\uparrow$ & FP $\downarrow$ & FN $\downarrow$ & TN $\uparrow$ \\
\hline
\multirow{8}{*}{Synthetic Dataset}
 & Traditional & JET & 0.53 & 0.90 & 0.50 & \textbf{50} & 942 & \textbf{0} & 4008 \\
\cline{3-10}
 & ML Baseline & Isolation Forest & 0.61 & 0.98 & 0.68 & \textbf{50} & 169 & \textbf{0} & 4781 \\
\cline{3-10}
 & \multirow{6}{*}{LLMs}
 & Gemma 2B & 0.53 & 0.92 & 0.53 & 49 & 685 & 1 & 4265 \\
 &  & Gemma 7B & 0.71 & \textbf{0.99} & 0.79 & \textbf{50} & 68 & \textbf{0} & 4882 \\
 &  & Llama 3.1 8B & 0.67 & 0.88 & 0.73 & 39 & 78 & 11 & 4872 \\
 &  & Mistral 8B & \textbf{0.90} & 0.98 & \textbf{0.94} & 48 & \textbf{12} & 2 & \textbf{4938} \\
 &  & Mistral Small 22B & 0.53 & 0.92 & 0.52 & 49 & 711 & 1 & 4239 \\
 &  & GPT-5-mini & 0.55 & 0.95 & 0.56 & \textbf{50} & 466 & \textbf{0} & 4484 \\
\hline
\end{tabular}
}
\end{table}

\vspace{-8pt}
\begin{table}[h]
\centering
\caption{Results on Anonymised Dataset, along with prompt ablations. Best values per column in \textbf{bold}. 
AuditCopilot is the prompt variant shown in Fig.~\ref{fig:prompt-vanilla}; 
\textit{w/o IF} removes Isolation Forest results; 
\textit{w/o Stats, IF} removes both global statistics and Isolation Forest hints.}
\label{private_ablation}
\vspace{-5pt}

\renewcommand{\arraystretch}{1.3}
\setlength{\tabcolsep}{6pt}

\resizebox{\textwidth}{!}{%
\begin{tabular}{lllcccccccc}
\toprule
Dataset & Prompt Variant & Method &
Precision~(\%) & Recall~(\%) & F1~(\%) &
TP~$\uparrow$ & FP~$\downarrow$ & FN~$\downarrow$ & TN~$\uparrow$ \\
\midrule

\multirow{11}{*}{Private Dataset}

& - & Isolation Forest
& 0.30 & 0.96 & 0.46
& 315 & 719 & 14 & 3952 \\

\cline{2-10}
& \multirow{5}{*}{AuditCopilot}

& Gemma-2B          
& 0.30 & \textbf{0.98} & 0.46
& \textbf{323} & 740 & \textbf{6} & 3931 \\

& & Mistral-8B        
& 0.39 & 0.90 & 0.54
& 295 & 468 & 34 & 4203 \\

& & Gemma-7B          
& \textbf{0.89} & 0.78 & \textbf{0.83}
& 256 & \textbf{32} & 73 & \textbf{4639} \\

& & Llama-3.1-8B      
& 0.17 & 0.86 & 0.28
& 282 & 1369 & 47 & 3302 \\

\cline{3-10}
& \multirow{3}{*}{w/o IF}

& Gemma-2B           
& 0.32 & 0.79 & 0.46
& 259 & 543 & 70 & 4128 \\

& & Gemma-7B          
& 0.14 & 0.90 & 0.24
& 311 & 1973 & 34 & 2982 \\

& & Llama-3.1-8B      
& 0.18 & 0.81 & 0.29
& 267 & 1232 & 62 & 3439 \\

\cline{3-10}
& \multirow{3}{*}{w/o Stats, IF}

& Gemma-2B           
& 0.07 & 0.07 & 0.07
& 23 & 290 & 306 & 4381 \\

& & Gemma-7B          
& 0.41 & 0.22 & 0.28
& 71 & 104 & 258 & 4567 \\

& & Llama-3.1-8B      
& 0.07 & 0.35 & 0.11
& 112 & 1668 & 217 & 3003 \\
\bottomrule
\end{tabular}
}
\end{table}

\section{Conclusion}
\vspace{-10pt}
Our results demonstrate that prompt-engineered LLMs, when combined with Isolation Forest scores, 
can outperform traditional JETs and ML baselines by offering both strong anomaly detection and 
natural-language rationales. While challenges remain around false positives, cost, and real-world 
deployment, this study highlights the promise of AI-augmented auditing where accuracy and inter-
pretability are jointly optimized. Importantly, we provide the first evidence that LLMs can be applied to tax-related ledger data, a 
high-stakes domain where transparency and trust are critical. Our findings suggest that combining 
LLMs with classical anomaly signals offers a practical path forward, balancing precision, recall, and 
explainability. Future work will extend these insights to larger, more diverse datasets and explore 
robustness against prompt variation and model updates.

\vspace{-10pt}
\section*{Limitations}
\vspace{-10pt}
\paragraph{Data scope and labels.} 
Our evaluation is limited to one anonymized tax-related ledger and a synthetic benchmark~\cite{pub15550}. Real-world ledgers differ across industries, ERP systems, and jurisdictions, and ground-truth labels are scarce. Pseudo-labels based on JETs may inherit rule-based biases, while synthetic anomalies are simulator-defined. Thus, external validity remains to be demonstrated with larger, multi-source datasets and independent audit labels.
\vspace{-10pt}
\paragraph{Model stability and explainability.} 
LLM behavior is prompt-sensitive and model-dependent: small changes in phrasing or sampling can shift results, and we observe disagreement across families (e.g., Gemma vs.\ Mistral). Although our AuditCopilot design reduces variance, stability under paraphrasing, adversarial inputs, and model updates is not guaranteed. Likewise, natural-language explanations may not always faithfully reflect decision boundaries, even if auditors find them useful.
\vspace{-10pt}
\paragraph{Deployment and governance.} 
While our method shows strong experimental performance, using LLMs in financial auditing raises safety, privacy, and compliance challenges. Risks include hallucinations, automation bias, prompt injection via free-text fields, and lack of reproducibility as model weights evolve. We stress that our approach should be viewed as \emph{decision support}, requiring human oversight, access controls, and governance frameworks before any real-world deployment.
\vspace{-10pt}
\section*{Acknowledgements}
\vspace{-10pt}
This work was supported by the German Federal Ministry of Education and Research (BMBF) under grant 01IW24006, and by the Federal Ministry of Research, Technology and Space under grant 16IS23064; it has also been supported by the Ministry for Science and Culture of Lower Saxony (MWK), the Endowed Chair of Applied Artificial Intelligence, Oldenburg University, and DFKI, with industry advisory support from DATEV eG.

\clearpage

\bibliography{refs}
\clearpage
\appendix

\section{Related Work}
\label{sec:related-work}

\paragraph{Machine Learning for Anomaly Detection.}  
A large body of research has explored machine learning (ML) to enhance anomaly detection in accounting data. Unsupervised algorithms, such as clustering and Isolation Forest, can sift through journal entries to pinpoint outliers that deviate from normal patterns~\cite{herreros2023applied}. On synthetic general ledger data, these methods improved detection beyond rule-based JETs; notably, Isolation Forest was found to be effective at uncovering fraudulent entries~\cite{pub15550}. To aid practitioners, explainability techniques like SHAP have been applied so that flagged anomalies are accompanied by insights about why they were deemed suspicious~\cite{herreros2023applied}. Still, such approaches are limited: traditional ML models operate mainly on structured features, produce risk scores or clusters, and often cannot provide natural-language justifications without additional tooling or domain expertise.

\paragraph{LLMs as Auditing Assistants.}  
The advent of Large Language Models (LLMs) has opened new avenues for auditing. These foundation models possess strong natural language reasoning abilities, making them promising co-pilots for auditors. For example,~\cite{gu2023aicopilot} fine-tuned a GPT-4 model with chain-of-thought prompting to support several audit tasks, including journal entry testing, and demonstrated how such a system can systematically guide audit procedures. Their work highlights the potential of LLMs to augment efficiency and provide richer insights in auditing contexts.

\paragraph{LLMs for Anomaly Detection.}  
Recent advances indicate that pretrained LLMs can also serve as anomaly detectors. SigLLM~\cite{mit2024sigllm} reformulates time-series sensor data into textual sequences and prompts an LLM to identify irregular patterns, achieving performance comparable to specialized algorithms without task-specific training. Similarly, AnoLLM~\cite{tsai2025anollm} serializes tabular records into text and leverages an LLM to assign anomaly scores, outperforming many conventional models across benchmark datasets. These approaches demonstrate the versatility of LLMs in ingesting heterogeneous data formats and applying contextual reasoning.

\paragraph{Auditing the Auditors.}  
Deploying LLMs in high-stakes audit settings raises concerns about consistency, bias, and hallucinations. To address this, the community has begun developing frameworks to evaluate and govern LLM-based auditors. AuditLLM~\cite{amirizaniani2024auditllm} probes model consistency across query rephrasings, while LLMAuditor~\cite{amirizaniani2024llmauditor} automates large-scale auditing with a combination of LLM checks and human-in-the-loop oversight. These works emphasize the importance of governance mechanisms alongside technical advances.

\section{Prompt Templates}
\label{app:review_prompt}

We used two standardized prompt formats: the \textit{Vanilla Prompt} for the private dataset 
(Figure~\ref{fig:prompt-vanilla}) and the \textit{Synthetic Prompt} for the synthetic dataset 
(Figure~\ref{fig:prompt-synthetic}).

\begin{figure*}[h]
\centering
\input{appendix_tables/prompt_template_vanilla}
\caption{AuditCopilot prompt template with dataset statistics and Isolation Forest hints}
\label{fig:prompt-vanilla}
\end{figure*}

\begin{figure*}[h]
\centering
\input{appendix_tables/prompt_template_synthetic}
\caption{Synthetic dataset prompt template with engineered flags and rule-based decision criteria.}
\label{fig:prompt-synthetic}
\end{figure*}

\end{document}